\documentclass[conference]{IEEEtran}

\usepackage{graphicx}

\title{\LARGE \bf
Autoencoder-augmented Neuroevolution for Visual Doom Playing
}
\author{Samuel Alvernaz and Julian Togelius\\
Department of Computer Science and Engineering\\
New York University, NY, USA\\
sja353@nyu.edu, julian@togelius.com
}

\begin{document}

\maketitle
\thispagestyle{empty}
\pagestyle{empty}

\begin{abstract}
Neuroevolution has proven effective at many reinforcement learning tasks, including tasks with incomplete information and delayed rewards, but does not seem to scale well to high-dimensional controller representations, which are needed for tasks where the input is raw pixel data. We propose a novel method where we train an autoencoder to create a comparatively low-dimensional representation of the environment observation, and then use CMA-ES to train neural network controllers acting on this input data. As the behavior of the agent changes the nature of the input data, the autoencoder training progresses throughout evolution. We test this method in the VizDoom environment built on the classic FPS Doom, where it performs well on a health-pack gathering task.
\end{abstract}

\section{Introduction}
When learning a policy for playing a game, it helps immensely if the representation of the game state (the input to the controller) is low-dimensional and structured in such a way that it can easily be processed by the controller. However, if the controller does not have access to the internal state of the game, it is forced to rely on the same information a human would have when playing the game, primarily the visual feed. This makes the learning task much harder, as we need to learn to transform the high-dimensional visual input to information that the controller can act on while learning the actual policy.

In general, working in three-dimensional environments, with solely visual input and no easy backend into a high-level representation of the game state, brings us closer to non-game (``real-world'') applications of these AI methods, such as robot control.
This is seen by many as a more relevant task for testing general intelligence in AI agents, which is why we have seen much work on learning from visual input recently.

So far, methods such as Deep-Q learning~\cite{MnihDeepQ} have seen success while working with large-scale visual input, while we have not seen successful applications of evolution to large-scale
visual problems, though it has been suggested that evolutionary methods may provide better results in partially-observable environments~\cite{togelius2009ontogenetic}. In other contexts, evolutionary algorithms have proven effective in generating interesting, unique strategies to succeed in difficult environments, and are capable of coming up with multiple solutions to the same problem; they have also performed well on tasks with delayed rewards and imperfect information~\cite{risi2015neuroevolution}. However, usually such algorithms require a pre-processed, high level representation of the environment, since raw visual data is too high-dimensional for evolutionary algorithms to effectively improve. (the larger the input, the larger the networks to process them must be, and the larger the networks, the larger the genome, and the larger the genome, the slower the evolution). Gradient-based deep learning methods, however, excel at compressing high-dimensional data and extracting relevant high-level, low-dimensional representations~\cite{compressionbasics}. Using these methods to create a compressed representation of the visual data of an environment, and then training a smaller network off of this compressed representation to generate agent behavior, could make it possible to enjoy the advantages of evolutionary algorithms when working with raw visual information. 

In this paper, an autoencoder - a particular network architecture described in further detail below - is used to create a compressed representation of a a three-dimensional video game environment, namely Doom. This compressed representation is then fed into a separate, smaller behavior-generating neural network, optimized via evolutionary methods, to demonstrate that the level of compression achieved by the autoencoder is sufficient to enable evolutionary optimization. It should be noted, however, that evolutionary methods are not necessarily vital to this method. For example, evolutionary optimization could be switched out for Deep-Q learning off of the compressed inputs. 

\subsection{Autoencoders}

An autoencoder is a network composed of three basic parts: The encoder, the chokepoint, and the decoder~\cite{Autoencoderfeatureextraction}. These networks are trained to reproduce their own input as their output, the idea being that the chokepoint will force the network to learn patterns in the data that allow it to be compressed to fit the chokepoint---because the decoder will have to reconstruct the input, after all, from all the information that something as small as the chokepoint can represent. 

In a videogame environment, we can usually depend on a certain uniformity in the environment; different frames look somewhat like each other (the environment will usually be constructed of similar elements, just viewed from different angles). If an autoencoder can successfully encode a single image frame from the environment, perhaps it could generate a diversity of representations of any image frame from the environment, so that it could represent the entire environment in a highly compressed manner. And if we're able to train an autoencoder to compress all frames from a video game environment (even if the decoder's reconstruction of these images is foggy and incomplete), then we could train a smaller network off of the autoencoder's chokepoint compression of these frames, and use it to generate behavior in the gameplay environment. If the resulting chokepoint was low-dimensional enough, we could use standard neuroevolutionary techniques to optimize the weights for this smaller network to play the game well.
\begin{figure}
\begin{center}
\includegraphics[width=\columnwidth]{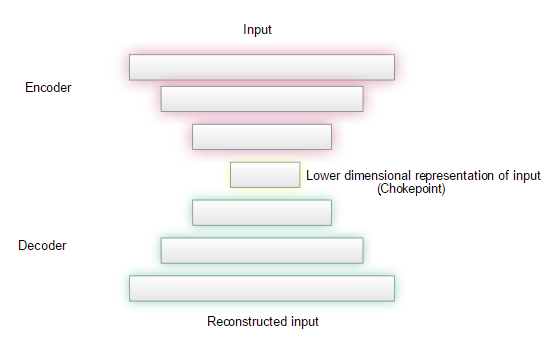}
\caption{Simple autoencoder diagram}
\end{center}
\end{figure}
A simple, high-level way to think of this is that the autoencoder will be converting the raw pixel data from game images into a high-level representation of the environment, like having a picture of how the world looks in your head. Even if the picture is fuzzy or incomplete, it's easier to make decisions based off that small picture rather than by simply considering the large arrays of raw visual information.

\begin{figure*}
\begin{center}
\includegraphics[width = \textwidth]{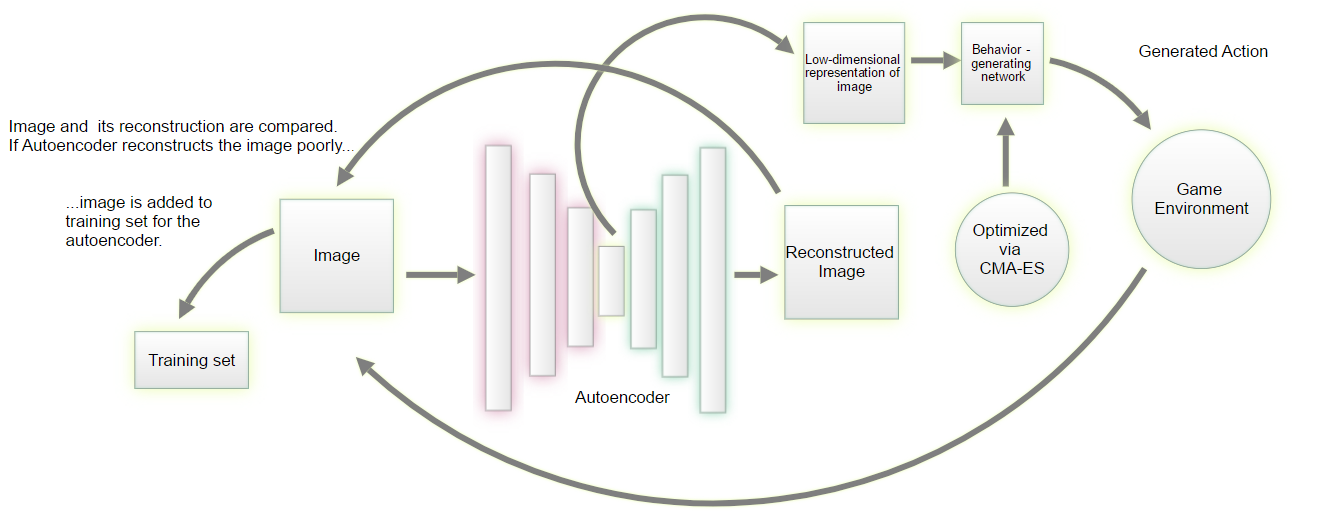}
\caption{Simple diagram of the method used in this paper. Image from the game environment is fed into the autoencoder. The autoencoder generates two outputs: The lower-dimensional representation of the image from the chokepoint, and the reconstructed image at the output layer. The lower-dimensional representation of the image is fed into the behavior-generating network, which generates an action in response to take within the game environment. Once the action step is taken in the game environment, a new image is generated, which is fed back into the autoencoder to generate the next action. In addition, each image input into the autoencoder is compared to its own reconstruction from the output layer. If the reconstruction of the input image isn't very good, the image is added to the training set for the autoencoder. Training of the autoencoder takes place between generations.}
\label{fig:Process Diagram}
\end{center}
\end{figure*}

\section{Previous work}

The idea of compressing visual input so that evolutionary algorithms may be used has been demonstrated before in fairly simple two-dimensional environments~\cite{MountainCar}. In addition, it has been previously shown that autoencoders can reduce high-dimensional visual input to a very small representation at its chokepoint, reducing a simple visual environment down to as small as two values representing the x and y position of the agent in the environment~\cite{Maron}. Autoencoder compression of the environment has been shown to be successful when combined with Q-learning, as well, for successful results in simple Atari environments before the current popularity of Deep-Q learning\cite{Lange}, as well as on simple real-world applications such a slot car control~\cite{lange2012autonomous}. Work has also been done in evolving neural network agents in a more complex 3-dimension FPS environments such as Quake by Parker et al~\cite{parker2009backpropagation}, though this was done by first preprocessing the raw visual input into a much smaller array of pixel values by averaging the value of pixels over a pre-defined block.

A similar method to the one presented in this paper was effective in the TORCS (The Open Racing Car Simulator) environment, demonstrated by Koutnik et al.~\cite{KoutnikTORCS}. However, rather than training their compressor through backpropagation (as in this paper), the weights for the compressing network were evolved to maximize diversity in the feature vector representation of the environment. Success was demonstrated in generating an agent that could remain on a track in a car-driving environment. 

While most work with vision-based AI training has taken place in simpler, two-dimensional environments, there have been forays into three-dimensional games, particularly in the Doom environment, using Deep-Q learning methods~\cite{deepQDoom,SLAMdoom,IntelDoom}. Common to all these approaches is that they supplement visual information about the environment with other variables taken from a backend into the game, whether it is information about health, ammo, a depth buffer, etc. While these methods have achieved impressive results, information such as this is not necessarily always something that would be available to a real-world controller.

Outside of the realm of generating agents to play in game environments, combining autoencoders with neuroevolution has been shown to generate interesting results with content creation by Liapis et al with their system DeLeNoX~\cite{liapis2013transforming}. Autoencoders were used to characterize two-dimensional spaceship designs as a low-dimensional array, which was used to refine the exploration of a possibility space for designs of believable-looking spaceships.  In addition, the hierarchal representational capabilities of classifier networks have been used in cooperation with neuroevolution to provide a measure of fitness when generating new, unique images, evolving networks to generate images that match the classifier's labels, resulting in interesting representations of various classes that are completely artificially generated~\cite{nguyen2015innovation}.

Evolution strategies are a class of evolutionary algorithms that are particularly well suited to optimization in real-valued space. The weights of a neural network can be seen as a real-valued vector, and optimized with evolution strategies or other similar algorithms; the training of neural networks with evolutionary algorithms is often called neuroevolution~\cite{floreano2008neuroevolution,risi2015neuroevolution}. Neuroevolution has long been known to be an effective reinforcement learning method, in particular when training comparatively small networks~\cite{igel2003neuroevolution,gomez2006efficient}. Methods for scaling up neuroevolution to apply them to larger networks have been developed, focusing on encoding or compressing the genome of the network in lower-dimensional space~\cite{koutnik2013evolving,stanley2009hypercube}. Recently, neuroevolution has also been shown to be a highly scalable alternative to more mainstream reinforcement learning methods when training networks for behavior in high-dimensional environments such as Atari games, achieving competitive results when compared to methods such as Deep-Q learning~\cite{Largescaleevolution}. In addition, it was demonstrated that evolution strategies can also be successful in constructing the topology of deep neural nets as well, with an extension of the popular NEAT~\cite{Stanley} (Neuroevolution of Augmenting Topologies) method into the realm of deep learning~\cite{CoDeepNeat}. 

This work distinguishes itself by taking raw visual input, with no preprocessing, and using an autoencoder trained via backpropagation to compress this input to a size more manageable for evolutionary methods. The training set of images for the autoencoder is obtained while playing the game with evolved agents that require no manual coding for the environment, minimizing the amount of necessary human input into the system. 

\section{Method}

Our experiments were carried out using \emph{VizDoom}~\cite{VizDoom}, a platform that supports agents in various environments based on the popular 1993 "Doom" video game. VizDoom was chosen because it is a popular implementation of the FPS Doom, a three-dimensional game that provides arrays of RGB visual information every frame, and which implements several different scenarios that concentrate on different goals, such as health-gathering, monster-killing, or navigation through the environment. The autoencoder was created using \emph{Keras}, a high-level deep learning package for python that can run off of either theano or tensorflow. Weights for the behavior-generating network were evolved using a python implementation of CMA-ES~\cite{CMAES,hansen2001completely} (Covariance Matrix Adaptation Evolution Strategy). CMA-ES was chosen as a popular, readily available evolutionary method with broad application, and the purpose of evolution was to demonstrate that the autoencoder had sufficiently compressed the visual input so that evolutionary optimization could be possible. Other evolutionary methods could be substituted in to evolve the behavior-generating network.

In this paper, experiments were run in the VizDoom "health gathering" environment. In this environment, the player is in a room that is filled with acid, slowly deteriorating their health, and must move about to gather health packs to prevent themselves from dying. Small red jars act as mines, and if picked up, will damage the player. The objective is to survive for as long as possible. Three actions are available: turn left, turn right, and move forward. There is a certain element of luck to the environment: health packs spawn randomly, and the player is spawned at random locations in the map. 

To generate behavior, an autoencoder is trained to reproduce input images of the environment (environment represented by 120x160x3 RGB images.) Table 1 shows the topology of the autoencoder network. All layers used ReLU (rectified linear unit) activation. ReLU is a simple activation function, where ReLU(x) = x if x is greater than 0, and 0 if x is less than zero. A small fixed-topology
network is then fed the input from the autoencoder's chokepoint (fc2 in the table, 128 float values). All layers in the behavior-generating network used sigmoid activation.

The behavior network was asked for a behavior decision every five frames of gameplay. In the final layer of the behavior network, the first three values were used to indicate which actions to take, with a sigmoid activation of greater than 0.5 being used to indicate that the action should be taken. The final value was multiplied by 5 and rounded up to the nearest integer to indicate how many times the action should be repeated. So, for example, if the network was asked to generate an action, and returned the array [$0.2$, $0.7$, $0.7$, $0.6$], it would repeat action 1 and action 2 (assuming that the first action, represented here by the value 0.2, was referred to as action 0) three times, and then do nothing for the remaining two frames before it was asked for a behavior decision again. There were 2208 weights total being optimized in the behavior-generating network. 
\begin{table}[h]

\label{Autoencoder}
\begin{center}
\begin{tabular}{|c||c||c||c||c||c|}
\hline
Layer & Filter dim. & Stride & Filters & Input & Output\\
\hline
conv1 & 8x8x3 & 4 & 32 & 120x160x3 &30x40x32\\
\hline
conv2 & 4x4 & 3 & 64 & 30x40x32 &10x14x64\\
\hline
conv3 & 4x4 & 3 & 64 &10x14x64&4x5x64\\
\hline 
fc1 & & & & 1280 & 512\\
\hline
fc2 & & & & 512 & 128\\
\hline 
fc3 & & & & 128 & 512\\
\hline
fc4 & & & & 512 & 1024\\
\hline
fc5 & & & & 1024 & 120x160x3\\
\hline
\end{tabular}
\end{center}
\caption{Topology of the autoencoder network. Each row represents a layer in the network. The outputs from one layer feed into the inputs of the next. The first layer is fed the 120x160x3 array of pixel information representing the game environment. \emph{fc} refers to 'fully connected' layers, \emph{conv} refers to two-dimensional convolutional layers.}
\end{table}

\begin{table}[h]
\label{behavior network}
\begin{center}
\begin{tabular}{|c||c||c|}
\hline
Layer & Input & Output\\
\hline
fc1 & 128 & 16\\
\hline
fc2 & 16 & 8\\
\hline
fc3 & 8 & 4\\
\hline
\end{tabular}
\end{center}
\caption{Topology of the behavior-generating network. Table follows similar conventions to Table 1. The first layer is fed the 128 float value representation of the environment from the chokepoint of the autoencoder network.}
\end{table}

The autoencoder network was trained through backpropagation on the images generated by gameplay. The behavior-generating network was trained by evolutionary methods, specifically CMA-ES, as mentioned previously. For the first 30 generations, the behavior-generating network was rewarded for maximizing novelty, where novelty was defined as seeing images that the autoencoder could not accurately reproduce, by comparing the reproduced images to the actual images taken fromt he environment. More specifically, it was calculated by taking the average of the absolute difference between the two image's pixel values. This was done to ensure that, through gameplay in the early stages of training, the images delivered to the autoencoder for training by backpropagation would be a diverse representation of the environment, training the autoencoder to reproduce as much of the environment as possible.  

After approximately 1,000,000 frames of gameplay, the autoencoder could reconstruct the environment to a sufficient degree that gameplay decisions could be made from the chokepoint values, so after 30 generations, the fitness function for the behavior-generating network was switched to reward it for actual fitness in gameplay (the number of frames the agent manages to survive in the hostile environment). The autoencoder continues to be trained by backpropagation along with the behavior-generating network after the first 30 generations. The training set for the autoencoder was generated from the actions the behavior-generating network took in the game environment. A simple filtering process was done on these images: Reconstructed images were compared to their input images (by calculating a simple absolute difference between their values) and if the average difference of the pixel values was below a threshold (.05) then the image was not included as part of the training set for the autoencoder to train on. Or, in other words, if the autoencoder already did a good job of reconstructing the input image, that image was excluded from the training set. 

\section{Environment reconstruction and performance results}
\begin{figure*}[h]
\begin{center}
\includegraphics[width = \textwidth]{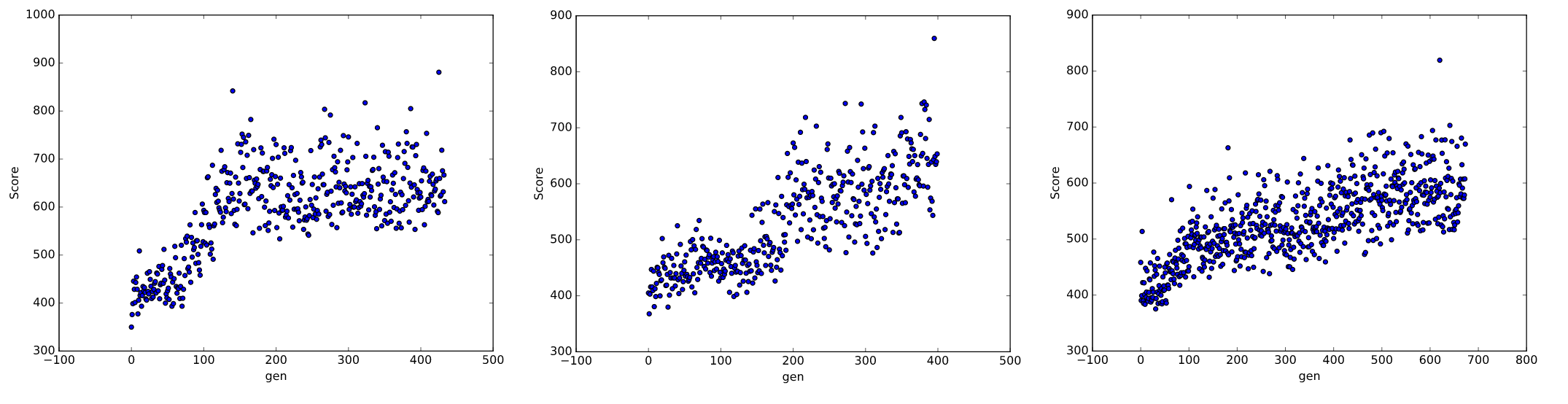}
\caption{Evolutionary progress of agents trained off of alternative autoencoder structure chokepoint output. Left: Network A. Center: Network B. Right: Network C.}
\label{fig:NetworksDE}
\end{center}
\end{figure*}

\subsection{Autoencoder compression of environment} 
Below are a couple of images demonstrating the quality achieved by the autoencoder in reconstructing the environment from the chokepoint input. Both of these images are taken from networks that have been trained for at least 300 generations.
\begin{figure}[h!]
\begin{center}
\includegraphics[width=\columnwidth]{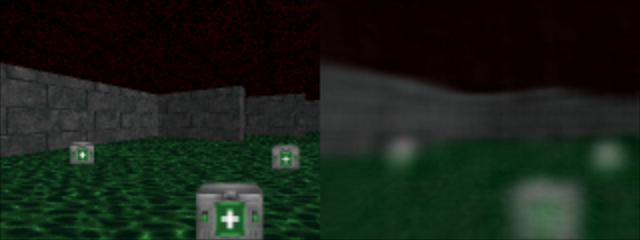}
\caption{Sample of autoencoder reconstruction of input image, generation 330}
\label{fig:normal medkit detection}
\end{center}
\end{figure}

In Figure~\ref{fig:normal medkit detection}, the original input image is on the left, and the autoencoder reconstruction of the input image is on the right. You can see the network does a decent job of reproducing the walls, floor, and multiple medkits at varying distances - so it would seem that the 128 floating points at the chokepoint can contain enough information to fairly accurately reproduce the images they are fed (composed of 57600 pixel values) even if the images are a bit fuzzy and ghostly. 
\begin{figure}[h!]
\begin{center}
\includegraphics[width=\columnwidth]{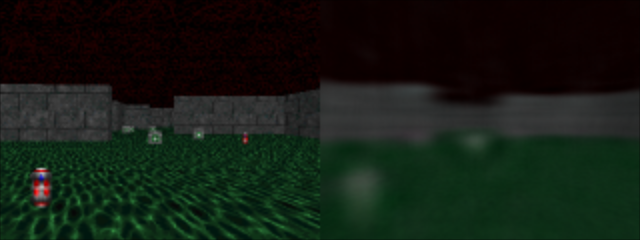}
\caption{Sample of autoencoder reconstruction of input image, generation 312}
\label{fig:normal mine detection}
\end{center}
\end{figure}

Figure~\ref{fig:normal mine detection} demonstrates a weakness of the autoencoder's reconstruction of the environment. It has difficulty properly reproducing the red jars. As mentioned above, the red jars act as a sort of mines - if they are picked up they damage the player. In this image, you can see the red mine closer to the player does seem to be detected by the network - but when reproducing it, it almost seems as if the network is trying to reproduce it as a health pack in the reconstructed image. And the red jar in the background, much further away, doesn't seem to be reproduced by the network at all, even when health packs the same distance away are. The simplest explanation for this is that the red jars are simply seen much less often than the health packs are, so the network has fewer examples to learn from to reproduce them properly. 

\subsection{Performance summary}
All of the following networks refer to behavior-generating networks, which use the CMA-ES evolutionary algorithm to optimize their weights. All of these networks (where applicable) used a compressed representation of the environment provided by the same autoencoder, which was trained via backpropagation. So what we have here are multiple behavior-generating networks that were trained off of input from the same autoencoder providing compressed representations of the environment.

A couple of different trials were run as a baseline. 

First, a network that was evolved off of input from the autoencoder (identical to Network A, described below) was fed random input. We will call this Network 1. (the idea being that if Network A showed no difference between being fed random input or input from the autoencoder, then obviously it was not successfully gaining useful information from the compressed input from the autoencoder.)

In addition, a behavior-generating network was evolved off random input. We will call this Network 2. (the idea being that if Network 2 reached the same level of performance while being evolved off of random input as networks that were evolved off of autoencoder input, then obviously the networks were not necessarily gaining any benefit by using the autoencoder input.)

There were a few different networks evolved off of autoencoder input: 

Network A, whose fitness function was a simple average of its score over 10 games, 

Network B, which was fed an additional piece of information in addition to the autoencoder's chokepoint compression - the player's current health, normalized over a value of 0-1. 

Network C, which operated in an environment where the red jars were much more deadly (usually resulting in instant death if they were picked up.) This was done to see if the autoencoder could actually distinguish between health kits and mines.

All scores in the below table were calculated over 1000 games, for networks evolved for 400 generations. 'Solved' is defined as a score of 2000 (at which point the environment times out.) 'Good' is defined as a score of 1000-2000. 'Mediocre' is defined as a score of 500-1000. 'Bad' is defined as a score of 0-500 (and is defined because this 'Bad' score basically falls into the range of what we'd expect from a random network.) The score is defined as the number of frames an agent survives within the environment.

\begin{table}[h]
\caption{Performance Results}
\label{performance}
\begin{center}
\begin{tabular}{|c||c||c||c||c||c||c|}
\hline
Network & Mean & std dev & Solved & Good & Mediocre & Bad\\
\hline
 1 & 376.16 & 151.95 & 0 & 1 & 113 & 886\\
\hline
 2 & 362.7 & 139.69 & 0 & 0 & 108 & 892\\
\hline
A & 515.97 & 288.8 & 1 & 64 & 328 & 607 \\
\hline
B  & 466.2  & 265.2 & 1 & 48 & 242 & 709\\
\hline
C & 563.3 & 329.8 & 3 & 74 & 360 & 563 \\
\hline
\end{tabular}
\end{center}
\end{table}
\section{Alternative Convolution}
\begin{figure*}[h!]
\begin{center}
\includegraphics[width = \textwidth]{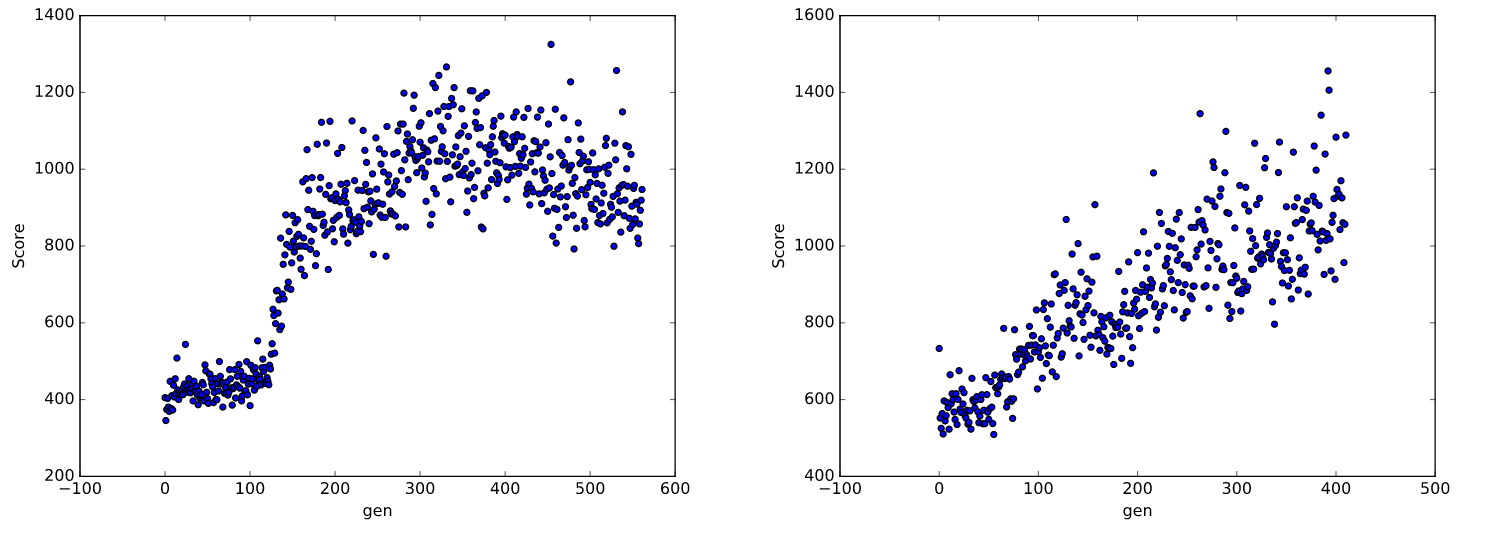}
\caption{Evolutionary progress of agents trained off of alternative autoencoder structure chokepoint output. Left: Network D. Right: Network E.}
\label{fig:NetworksDE}
\end{center}
\end{figure*}

While training the autoencoder to compress the environment, an alternative topology was accidentally stumbled upon that yielded some interesting results.

As described in Section 3, in the first layer, 8x8x3 filters were used on the image on the first layer. This is typically how 2D convolutional nets are used on color images - convolved over the image, looking for patterns in all three RGB channels. 

During a couple of the experiments being run, the images were fed into the autoencoder incorrectly. Specifically, the width value was put in place of the channel value. The specific effect this had was, instead of the first convolutional layer being constructed from 8x8x3 filters an 8x8 filter over the 'depth' of the three RGB channels), it was constructed from 3x8x160 filters (so each filter was looking for patterns in a very large space of 3840 pixel values. In other words, the width value was being treated as the channel value, so we had 3x8 size filters convolved over a 'depth' of 160 pixels.)

\begin{table}[h]
\caption{Autoencoder alternative Structure}
\label{Autoencoder}
\begin{center}
\begin{tabular}{|c||c||c||c||c||c|}
\hline
Layer & Filter dim. & Stride & Filters & Input & Output\\
\hline
conv1 & 3x8x160 & 4 & 64 & 120x160x3 &1x30x64\\
\hline
conv2 & 1x4 & 2 & 128 & 1x30x64 &1x15x128\\
\hline
conv3 & 1x4 & 2 & 256 &1x15x128 &1x8x256\\
\hline 
fc1 & & & & 2048 & 512\\
\hline
fc2 & & & & 512 & 128\\
\hline 
fc3 & & & & 128 & 512\\
\hline
fc4 & & & & 512 & 1024\\
\hline
fc5 & & & & 1024 & 120x160x3\\
\hline
\end{tabular}
\end{center}
\end{table}

What was interesting about this mistake was not only did this odd network architecture manage to reconstruct the environment fairly well, it actually seemed to give better results when it came to evolving behavior networks. 

\subsection{Alternative Autoencoder Compression of the Environment}

\begin{figure}[h!]
\begin{center}
\includegraphics[width=\columnwidth]{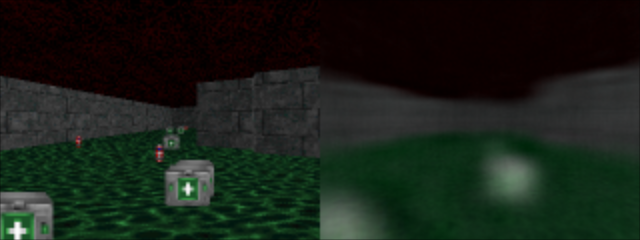}
\caption{Sample of alternative autoencoder reconstruction of input image, generation 374}
\label{fig:exotic medkit detection}
\end{center}
\end{figure}

Here, in Figure~\ref{fig:exotic medkit detection}, we see the alternative autoencoder's reconstruction of the gameplay environment, including a couple of health packs at varying distances. Again, original image is on the left, reconstruction is on the right. Compare to figure 2: we can see that this alternative structure is capable capturing the walls, floor, and the health packs. However, the reconstruction of the health packs is a bit less detailed than they are in the normal autoencoder's reconstruction in Figure~\ref{fig:normal medkit detection}, mostly being presented as white blobs. Nonetheless, they're still there, so the behavior generating network should be able to make decisions taking them into account. 

\begin{figure}[h!]
\begin{center}
\includegraphics[width=\columnwidth]{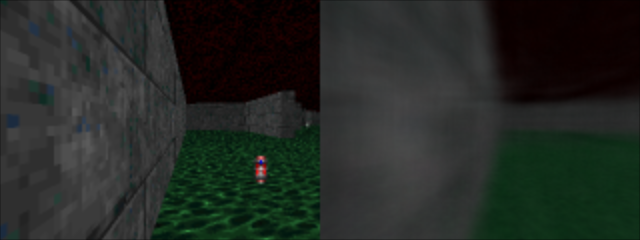}
\caption{Sample of alternative autoencoder reconstruction of input image, generation 418}
\label{fig:exotic mine detection}
\end{center}
\end{figure}

On the other hand, in Figure~\ref{fig:exotic mine detection}, we can see that the alternative autoencoder structure actually fails to reproduce the red jars at all. Generally, this particular network topology could not reproduce the red jars unless they were very close to the screen, and then only very faintly. This could have implications for performance - if the network cannot reproduce them, they may not be reflected at all in the chokepoint compression of the environment, and the network may not be able to make decisions taking their existence into account.

\subsection{Performance Summary}

As with the normal autoencoder structure detailed in section IV, Network 1 in the table below is a baseline network, detailing the results of a behavior generating network that was trained on autoencoder input when it was fed random input. Network 2 is the results of a network trained to optimize on random input. Network D is a behavior generating network trained on top of the alternative autoencoder compression whose fitness function is average performance over 10 games. Because Network D had such a high standard deviation, Network E was also trained, whose fitness function can be described by the following equation:

$$
5\alpha/3 - \beta 
$$

Where alpha is the network's score over 10 games, and beta is the standard deviation over those 10 games (so the network was punished for inconsistent performance).

\begin{table}[h]
\caption{Alternative Performance Results}
\label{altperformance}
\begin{center}
\begin{tabular}{|c||c||c||c||c||c||c|}
\hline
Network & Mean & std. dev & Solved & Good & Mediocre & Bad\\
\hline
1 & 369.19 & 147.93 & 0 & 0 & 110 & 890\\
\hline 
2 & 364.2 & 137.97 & 0 & 1 & 103 & 896 \\
\hline
D & 646.58 & 397.93 & 12 & 162 & 360 & 466\\
\hline
E & 657.1 & 397.05 & 12 & 170 & 374 & 444\\
\hline
\end{tabular}
\end{center}
\end{table}

\section{Video Results and Behavior Discussion}
\begin{figure}[h!]
\begin{center}
\includegraphics[width=\columnwidth]{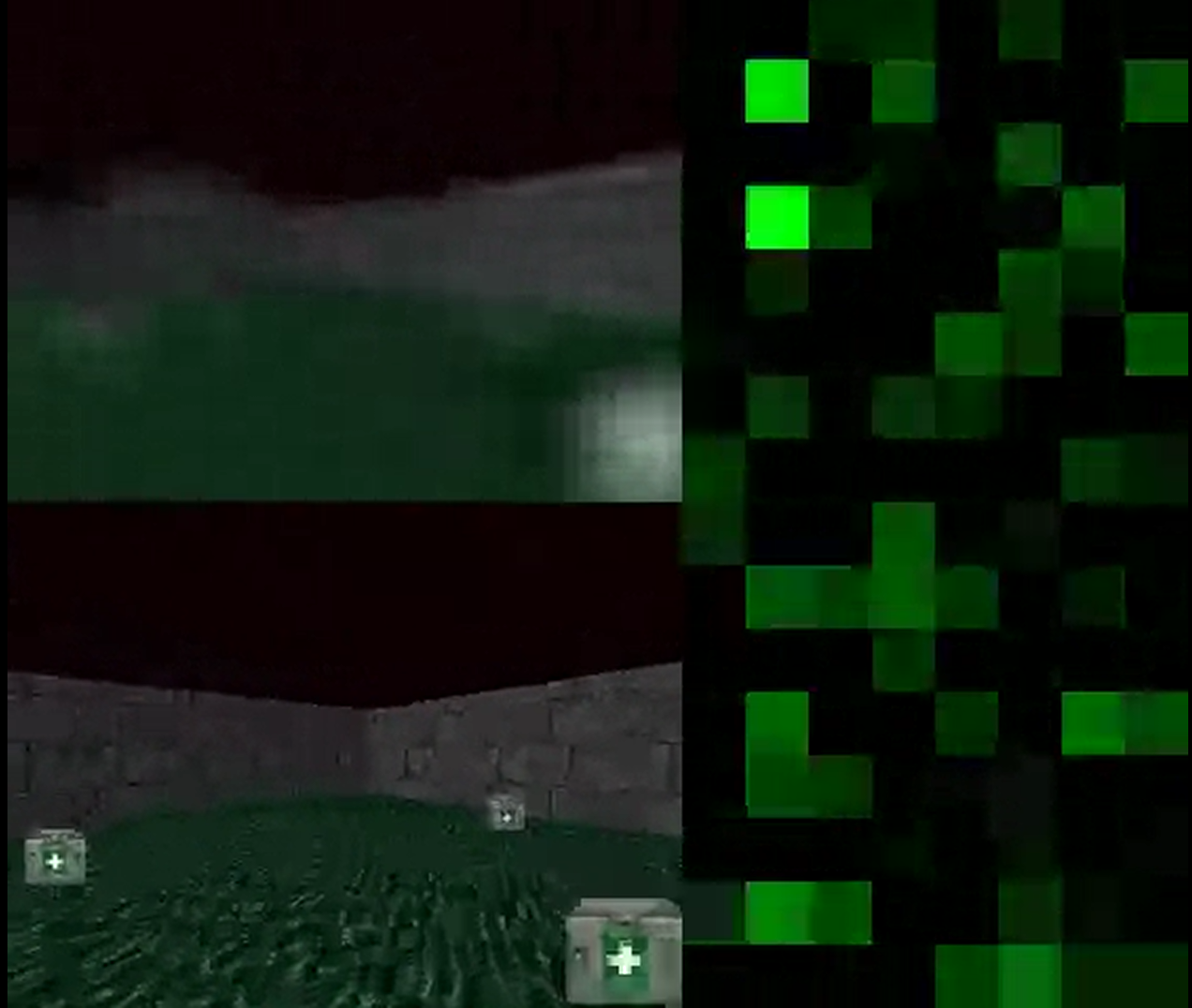}
\caption{Clip from video of generated agent play. Bottom left: Actual gameplay footage. Top left: autoencoder reconstruction of the environment. Right: Visual representation of chokepoint activation for this frame.}
\label{fig:video clip}
\end{center}
\end{figure}

Figure~\ref{fig:video clip} is a video clip visualizing the network behavior\footnote{Videos can be found here: https://tinyurl.com/m4df7rn}.

The evolution of networks A and B generated mostly disappointing behavior. While they do manage to navigate and pick up health packs, they evolved to only be able to turn left and right, respectively. This means that sometimes, they get stuck in uninteresting loops around empty rooms and die. Network B is interesting in that its information about the current state of its health seems to provide it an ability to break out of these loops. Networks C, D and E, however, were much more interesting. They all demonstrated an ability to navigate the environment and seek out health packs, to varying degrees of competence. Network C, by being trained in an environment where picking up a mine could immediately kill it, may have been bumped out of a local maxima that A and B fell into. 

But why did networks trained on the alternative autoencoder topology get better results, and much faster, than networks trained on an autoencoder doing typical convolution? If we compare the chokepoint activation of the two networks, we can see that the alternative autoencoder had a much sparser activation. Taking the sum of every activation of the chokepoint (so, for an individual frame, the maximum activation would be 128), we see that normal convolution yields an average chokepoint activation sum of 24.3 over 1000 frames of gameplay, while the average sum for the alternative convolution is 12.7. Information about the environment, then, was compressed into fewer values. While the model generated by the unusual convolution wasn't as accurate as the one generated by normal convolution (the red jars in the environment were almost invisible to it) the higher compression allowed evolution to take place at a much faster pace. This suggests that, perhaps, the environment could be compressed even further by normal convolution, down to something less than 128 values. This could aid evolutionary methods by necessitating smaller behavior-generating networks. 

\section{Conclusion}
Deep-Q learning off of the raw visual data in this environment outperforms the method presented here, learning to achieve a score between 1000-1500 after 500,000 steps of gameplay~\cite{VizDoom}. However, while it would have been nice to create a method that outperformed Deep-Q, the focus of this paper was on demonstrating that the hierarchal representation capabilities of deep-learning networks could compress the representation of raw visual data in a 3d environment to the degree that evolutionary methods could be effective. Such a compressed representation could be useful in other contexts: as suggested before, Deep-Q learning could be trained off of the compressed inputs as well, and perhaps training off of the compressed data would yield some advantage over training off of the raw visual data. Having a low-dimensional, internal representation of the environment, however fuzzy, could also enable other training methods beyond Deep-Q.

While the results from evolution in this paper do not demonstrate a mastery of learning strategy for playing in a three dimensional environment, it has been shown that autoencoder compression can build an internal, low-dimensional model of a three-dimensional FPS environment from the high-dimensional visual data. Moreover, it is not necessary to have a preset database comprising a balanced representation of the environment, as the autoencoder can be trained as the game is played. Evolutionary methods can be used on this compressed representation to improve agent behavior. 

\section*{Acknowledgements}
Thanks to to Christopher Dimauro for keeping the machines running and to Gabriella Barros and Christoph Salge for their advice. We also thank Nvidia for their generous hardware donation to the NYU Game Innovation Lab which made this project possible.

\nocite{*}
\bibliography{mybib.bib}{}
\bibliographystyle{plain}
\end{document}